%% file: main.tex
\documentclass{article}

\usepackage[preprint]{conference_2026}

\usepackage[utf8]{inputenc}
\usepackage[T1]{fontenc}
\usepackage{hyperref}
\usepackage{url}
\usepackage{booktabs}
\usepackage{amsfonts}
\usepackage{amsmath}
\usepackage{amssymb}
\usepackage{nicefrac}
\usepackage{microtype}
\usepackage{xcolor}
\usepackage{graphicx}
\usepackage{array}
\usepackage{multirow}
\usepackage{makecell}

\graphicspath{{figures/}}

\title{Cohort-Scale Neural Atlases of Ultrasound Video}

\author{%
  Zhuorui Zhang$^{1}$ \quad Roger Pallar\`es-L\'opez$^{1}$ \quad Xuan Wu$^{1}$ \\ \textbf{Praneeth Namburi$^{2,3}$  \quad Brian W. Anthony$^{1,2,3}$  \thanks{Corresponding author.}} \\
  $^{1}$Department of Mechanical Engineering, MIT \\
  $^{2}$Institute for Medical Engineering and Science, MIT \\
  $^{3}$MIT.nano Immersion Lab, MIT \\
  \texttt{\{zhuoruiz,roger31,xuanwu,praneeth,banthony\}@mit.edu}
}

\begin{document}

\maketitle

\begin{abstract}
Ultrasound is the most widely used real-time imaging modality in clinical practice, yet per-frame video annotation remains a major bottleneck: expert labels are scarce and costly, and image appearance varies with speckle, shadowing, attenuation, and operator-dependent probe pose. This is especially limiting because clinically relevant information is often dynamic, from left-ventricular motion in echocardiography to muscle and bone kinematics in musculoskeletal imaging. Population atlases can amortize annotation cost by registering observations to a shared canonical coordinate system, but existing neural atlas methods mainly target single videos, small test-time image sets, or object-centric image collections. We introduce a cohort-scale neural atlas for ultrasound video: a single canonical chart with per-video Generative Latent Optimization embeddings, trained jointly over thousands of frames in DINOv3 feature space. Across five cardiac and musculoskeletal datasets with point landmarks and segmentation masks, our method learns coherent canonical templates and enables accurate atlas-space annotation transfer. On EchoNet-Dynamic and MSK-Bone, it supports single- and few-shot transfer with accuracy competitive with strong dense-correspondence baselines, while training in minutes on a single consumer GPU. The learned embeddings are interpretable: linear projections reveal structured cohort variation, image-decoder interpolation produces anatomically plausible intermediate frames, and test-time latent inversion reconstructs held-out frames through the atlas. These results suggest that cohort-scale neural atlases offer a practical, interpretable representation for reducing expert annotation burden in ultrasound video analysis.
\end{abstract}

\section{Introduction}
\label{sec:intro}

Ultrasound represents the most prevalent real-time imaging modality in clinical workflows:
it is portable, radiation-free, and available at the bedside. For this reason, its clinical value usually depends not on isolated frames 
but on the motion patterns they reveal. In echocardiography, ejection fraction and chamber
kinematics are read from the moving left-ventricular silhouette across
the cardiac cycle~\citep{ouyang2020echonet,deng2024memsam}. In musculoskeletal (MSK) scanning, ultrasound makes it possible to observe internal muscle and connective-tissue dynamics during movement, revealing tissue motion patterns that are not apparent from external limb kinematics alone~\citep{namburi2025efficient}. However, quantifying these signals 
at scale requires dense, temporally consistent correspondence across frames, subjects, and acquisitions.

Three properties make dense correspondence in ultrasound especially difficult.
\textbf{(i)} Ultrasound appearance is dominated by multiplicative speckle,
shadowing, attenuation, and artifacts that vary with acoustic path and probe
pose, making pixel-level appearance highly non-stationary even within a single
clip~\citep{deng2024memsam}.
\textbf{(ii)} Acquisition is strongly operator-dependent: probe orientation,
contact pressure, gain, depth, and view selection induce appearance shifts
across sonographers, sites, and visits, often entangling anatomy with acquisition
variability.
\textbf{(iii)} Expert annotation is slow, costly, and scarce. Annotating
EchoNet-Dynamic~\citep{ouyang2020echonet} at standard clinical temporal density
would require nearly two thousand clinician hours, so clinicians typically label
only key frames such as end-systole and end-diastole~\citep{deng2024memsam}.
This burden grows further for larger private cohorts.
Together, these factors make fully supervised natural-video paradigms poorly
suited to ultrasound. Yet clinically relevant tasks such as longitudinal
tracking, biomarker estimation, phenotype discovery, and acquisition quality
control require dense, reliable correspondence across frames, subjects, studies,
and sites.

\begin{figure}[t]
  \centering
  \includegraphics[width=1.0\linewidth, trim={1pt 11pt 1pt 7pt}, clip]{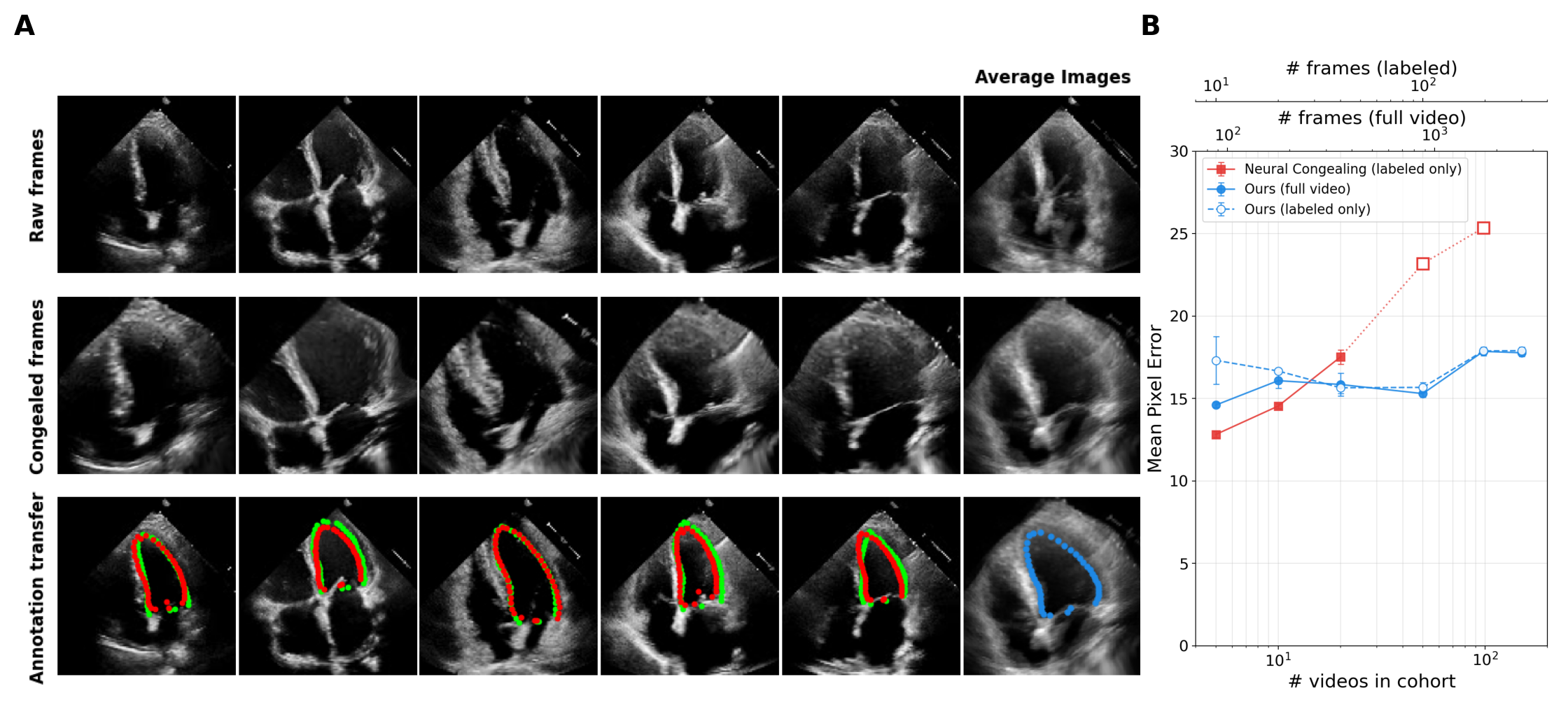}
  \caption{A cohort-scale EchoNet-Dynamic atlas supports few-shot annotation
    transfer and scales beyond 100 videos and 1000 frames.
    \textbf{(A)} Raw frames are warped into a shared atlas coordinate system,
    where their average reveals cross-video alignment. Atlas-space source points
    (\textcolor{blue}{blue}) transfer accurately to individual frames, matching
    ground truth (\textcolor{green!70!black}{green}) with predictions
    (\textcolor{red}{red}).
    \textbf{(B)} Mean pixel error versus cohort size on EchoNet-Dynamic subsets
    ($256\times256$ canvas; lower is better; error bars: $\pm$1 std over 3 seeds).
    Our full-video and labeled-only atlas variants remain stable across cohort
    sizes, whereas Neural Congealing~\citep{ofriamar2023neural} fails to converge
    for $\geq100$ frames; dashed red boxes report the best MPE before divergence.}
  \label{fig:hero}
\end{figure}

Population atlases provide a natural mechanism for amortizing annotation:
once anatomy is represented in shared coordinates, labels or measurements
defined in atlas space can be transferred to any subject that registers to the
atlas~\citep{cabezas2011atlasseg}. Classical medical-imaging pipelines such as
FreeSurfer~\citep{fischl2012freesurfer} and ANTs/SyN~\citep{avants2008syn}
operationalize this idea around canonical
templates~\citep{talairach1988atlas,evans1993mni,collins1994mni}, but
ultrasound has largely resisted atlas-style alignment because of speckle,
view variability, and operator dependence. Recent neural atlas and congealing
methods learn canonical representations for single videos, small test-time
image sets, or object-centric image
collections~\citep{kasten2021layered,ouyang2024codef,peebles2022gangealing,ofriamar2023neural,gupta2023asic,zhang2024congealing3d}.
However, they do not learn a single reusable atlas jointly over hundreds of
ultrasound videos and thousands of frames for cohort-scale annotation transfer.

We formulate cohort-scale atlas learning in analogy with neural radiance
fields~\citep{mildenhall2020nerf}: a shared canonical representation is observed
through video-specific latent variables that account for variation across
subjects and acquisitions. In ultrasound, however, directly learning this
representation in pixel space would entangle anatomy with scan-specific
appearance effects. We therefore learn the atlas in DINOv3 feature
space~\citep{simeoni2025dinov3}, where representations are more stable for
anatomical alignment and better suited for transferring annotations across
videos.

Concretely, we learn a shared canonical chart and per-video
Generative-Latent-Optimization embeddings~\citep{glo} (GLO), optimized
jointly across hundreds of videos (thousands of frames) in DINOv3
feature space. The resulting model is compact,
trains in minutes on a single RTX 4090, and exposes a structured
latent representation that can be queried for annotation transfer and
visualized through an external feature decoder.

We instantiate the atlas on five ultrasound datasets spanning cardiac
and musculoskeletal anatomy. The same architecture and training recipe
produce coherent canonical alignment across all five
(Fig.~\ref{fig:datasets-hero}), and we benchmark few-shot transfer in
detail on EchoNet-Dynamic (cardiac) and MSK-Bone (musculoskeletal). Specifically, the paper makes three contributions:

\begin{itemize}\setlength{\itemsep}{2pt}
\item \textbf{Cohort-scale neural atlas learning.} We train a
single canonical chart with per-video GLO embeddings at scale in foundation-feature space, using a compact four-head SIREN and a NeRF-W-style
heteroscedastic objective (\S\ref{sec:method}).
The model trains in minutes on a single RTX 4090.
\item \textbf{Few-shot annotation transfer across anatomies and
annotation types.} The same architecture produces coherent
atlas-space alignment on five ultrasound datasets covering cardiac and
musculoskeletal anatomy, point landmarks, surface-tracking landmarks,
and segmentation masks (Fig.~\ref{fig:datasets-hero}). On
EchoNet-Dynamic and MSK-Bone, the atlas achieves competitive single-
and few-shot transfer accuracy against baselines (\S\ref{sec:matching}).
\item \textbf{An interpretable atlas latent space.} Per-video embeddings show
structured variation under linear projection. Bilinear interpolation produces
anatomically plausible intermediate frames when decoded through an off-the-shelf
DINOv3-trained image decoder~\citep{shi2025svgt2i}. Latent inversion recovers
held-out frames by optimizing only a new embedding (\S\ref{sec:latent}).
\end{itemize}

\section{Related Work}
\label{sec:related}

\paragraph{Population atlases and medical registration.}
Medical image analysis has long relied on atlas-based registration:
stereotaxic and average-template construction~\citep{talairach1988atlas,
evans1993mni,collins1994mni}, diffeomorphic
registration~\citep{avants2008syn}, and atlas-warped
segmentation~\citep{cabezas2011atlasseg}. Learning-based methods such as
VoxelMorph~\citep{balakrishnan2019voxelmorph} replace iterative optimization
with feed-forward prediction, but still align single image pairs. Shared
canonical frames, when used, are typically fixed by expert anatomy or averaged
classical registrations. Our work keeps the population-atlas spirit but builds the atlas \emph{jointly with} the per-video mappings, in foundation-feature space, at cohort scale.

\paragraph{Neural atlases and congealing.}
Layered Neural Atlases~\citep{kasten2021layered} (LNA) and
CoDeF~\citep{ouyang2024codef} learn single-video canonical fields using
coordinate MLPs and hash-based deformation fields. GANgealing~\citep{peebles2022gangealing}
aligns image domains using a Spatial Transformer trained from a pretrained
StyleGAN2 generator, while Neural Congealing~\citep{ofriamar2023neural}
optimizes a DINO-ViT feature atlas for each small test-time image set.
ASIC~\citep{gupta2023asic} and 3D
Congealing~\citep{zhang2024congealing3d} extend canonical alignment to sparse
in-the-wild or 3D object-centric collections, building on classic
congealing~\citep{learnedmiller2006congealing}. Our work shifts the unit of
optimization from a single video or object/image set to a medical video cohort:
one atlas is trained jointly across many videos and reused for few-shot
annotation transfer.

\noindent\textbf{Implicit Neural Representations and Uncertainty.}
Implicit neural representations provide compact, resolution-independent mappings
from continuous coordinates to images, videos, geometry, and radiance
fields~\citep{tancik2020fourier,xie2021neuralfields,mildenhall2020nerf}.
SIREN~\citep{sitzmann2020siren} uses sinusoidal activations to model
high-frequency signals with smooth derivatives. NeRF in the
Wild~\citep{nerfw} and GLO~\citep{glo} introduced per-instance embeddings,
with NeRF-W also modeling heteroscedastic uncertainty for image-dependent and
transient effects. We combine these ingredients for cohort-scale ultrasound
atlas learning.

\noindent\textbf{Vision Foundation Model Representations.}
Self-supervised vision foundation models such as DINO~\citep{caron2021emerging},
DINOv2~\citep{oquab2023dinov2}, and DINOv3~\citep{simeoni2025dinov3} provide dense
features with strong semantic and correspondence structure that are more stable
than raw pixels under acquisition-dependent appearance variation. However, ViT
features remain spatially coarse and are not explicitly temporally smooth.
FeatUp~\citep{fu2024featup} addresses the spatial resolution gap with
coordinate-based feature fields for individual images. We instead use frozen
DINOv3 features to supervise a continuous spatiotemporal atlas for dense
correspondence and annotation transfer across ultrasound video cohorts.

\section{Method}
\label{sec:method}

\subsection{Setup and notation}
\label{sec:method-setup}

We are given a cohort $\mathcal{V}=\{\mathcal{V}_i\}_{i=1}^N$ of $N$ videos
sharing a common anatomical scene. Frames are indexed by normalized image coordinates
$(x,y)\in[-1,1]^2$ and per-video normalized time $t\in[-1,1]$. Each video
has a learned identity embedding $e_i\in\mathbb{R}^{d_e}$, and the slot
$e=\mathbf{0}$ is reserved as the canonical identity. A frozen vision foundation
model $\Phi$ produces patch features $F_i(x,y,t)\in\mathbb{R}^D$ that are
cached once. We default to DINOv3~\citep{simeoni2025dinov3} ViT-B/16 for
the per-dataset paired and few-shot evaluation atlases
(\S\ref{sec:matching}). Earlier cohort-scale demonstrations and the
latent-interpolation experiment (\S\ref{sec:latent-interp}) use ViT-S/16+.
Architecture and training are featurizer-agnostic..

\begin{figure}[t]
  \centering
  \includegraphics[width=\linewidth]{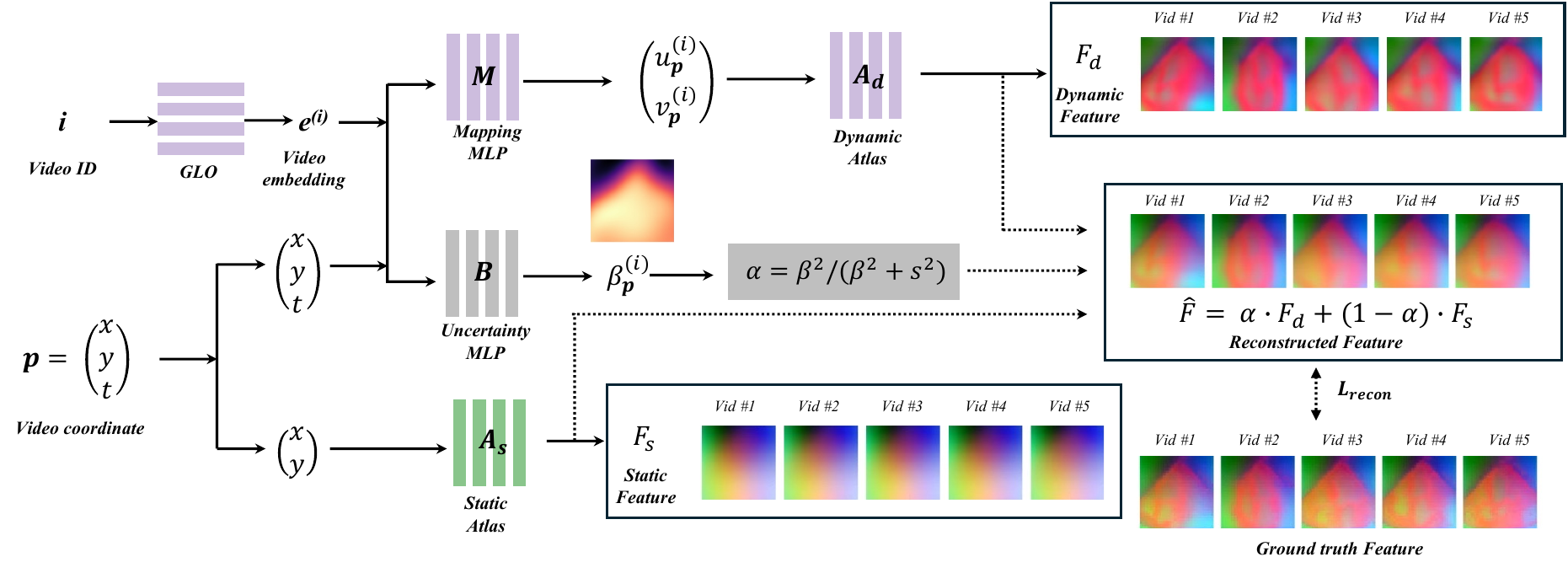}
  \caption{Cohort-scale four-head SIREN architecture. Each
  $(x,y,t,e_i)$ tuple is routed through three SIREN heads: the
  \emph{mapping MLP} $M$ deforms the coordinate into the canonical
  chart and the \emph{dynamic atlas} $A_d$ reads off shared,
  per-video-deformable anatomy. A small \emph{static atlas} $A_s$
  produces a slowly-varying cohort-shared prior. The
  \emph{uncertainty MLP} $B$ produces a per-pixel
  $\beta\!=\!B(x,y,t,e_i)$, whose monotone transform
  $\alpha=\beta^2/(\beta^2+s^2)$ blends the two paths into the
  composite $\hat F = \alpha F_d + (1-\alpha) F_s$.}
  \label{fig:arch}
\end{figure}

\subsection{Architecture}
\label{sec:method-arch}

The model has four SIREN heads with sinusoidal activations. The
\textbf{mapping MLP} $M:(x,y,t,e_i)\!\mapsto\!(u,v)$ maps
spatiotemporal coordinates from video $i$ into a canonical chart.
The \textbf{dynamic atlas} $A_d:(u,v)\!\mapsto\!\mathbb{R}^D$ stores
cohort-shared anatomy in this canonical space, while the
\textbf{static atlas} $A_s:(x,y)\!\mapsto\!\mathbb{R}^D$ is
intentionally low-capacity, and is mainly reserved for black padding outside the field of view that are common in ultrasound recordings, so that anatomical variation is represented
by the mapping and dynamic atlas rather than absorbed by a
high-capacity static field. The \textbf{uncertainty MLP}
$B:(x,y,t,e_i)\!\mapsto\!\mathbb{R}_{>0}$ produces
$\beta\!=\!B(x,y,t,e_i)$, which is transformed monotonically into
$\alpha\!=\!\beta^2/(\beta^2+s^2)$ with uncertainty scale $s$. The final feature
prediction is the uncertainty-weighted composition
\begin{equation}
\hat F(x,y,t,e_i) = \alpha\,A_d\!\big(M(x,y,t,e_i)\big)
                    + (1-\alpha)\,A_s(x,y).
\label{eq:composite}
\end{equation}

\subsection{Training objective}
\label{sec:method-objective}

Training minimizes
$\mathcal{L} = \mathcal{L}_{\mathrm{recon}}
+ \lambda_{\mathrm{anc}}\mathcal{L}_{\mathrm{anc}}
+ \lambda_{\mathrm{s}}\mathcal{L}_{\mathrm{s}}
+ \lambda_{\mathrm{d}}\mathcal{L}_{\mathrm{d}}
+ \lambda_{\mathrm{div}}\mathcal{L}_{\mathrm{div}}
+ \lambda_{\mathrm{rig}}\mathcal{L}_{\mathrm{rig}}
+ \lambda_{\mathrm{emb}}\|e\|_2^2.$
The \textbf{reconstruction} term is a NeRF-W heteroscedastic
NLL~\citep{nerfw},
$\mathcal{L}_{\mathrm{recon}}\!=\!\mathbb{E}[\|\hat F-F\|^2/\beta^2 + 0.5\log\beta^2]$,
which encourages $\alpha$ to emphasize the dynamic atlas where it
explains anatomy and to down-weight regions better modeled by the
static atlas. The \textbf{anchor}
$\mathcal{L}_{\mathrm{anc}}\!=\!\mathbb{E}\|M(x,y,t,e\!=\!\mathbf{0})-(x,y)\|^2$
fixes the canonical chart at the reserved $e=\mathbf{0}$ slot.
\textbf{Smoothness} terms for static atlas $\mathcal{L}_{\mathrm{s}}$, and dynamic atlas $\mathcal{L}_{\mathrm{d}}$ are asymmetrically weighted ($\lambda_{\mathrm{s}}$, $\lambda_{\mathrm{d}}$). \textbf{Diversity} term $\mathcal{L}_{\mathrm{div}}$ penalizes redundant
static/dynamic content. \textbf{Rigidity}
$\mathcal{L}_{\mathrm{rig}}\!=\!\mathbb{E}\|J_M J_M^\top\!-\!I\|_F^2$
encourages locally rigid deformations, where $J_M$ is the spatial Jacobian matrix of the mapping $M$ defined as:

$$
J_M =
\begin{bmatrix}
\dfrac{\partial u}{\partial x} & \dfrac{\partial u}{\partial y} \\
\dfrac{\partial v}{\partial x} & \dfrac{\partial v}{\partial y}
\end{bmatrix}.
$$

Optimization uses Adam with
two parameter groups (a higher learning rate for embeddings, a lower
one for all other parameters). 

\subsection{Single-shot and few-shot annotation transfer}
\label{sec:method-transfer}

Given source landmarks $\{(x_k,y_k)\}_{k=1}^M$ from video $a$ at time
$t_a$, we transfer them to target video $b$ at time $t_b$ in three
steps. First, each source point is mapped into atlas coordinates,
$uv_k\!=\!M(x_k,y_k,t_a,e_a)$. Second, in the few-shot
setting, corresponding atlas coordinates are averaged across support
videos. Atlas coordinates are used as is in single-shot (paired) setting. Third, the averaged atlas coordinates are inverted in the
target chart firstly by a searching algorithm that aims to minimize the error on the atlas coordinate. Specifically, we uniformly sample a 64$\times$64 grid for $x$ and $y$ each in range of [-1, 1], and then pick the best point with minimal projection error in atlas coordinate. Then we refine the searched points through gradient descent by
200 Adam steps that minimize the projection error as well. 

\subsection{Decoder for grayscale visualization}
\label{sec:method-decoder}

For the interpretability experiments in \S\ref{sec:latent}, we
visualize atlas composites as grayscale ultrasound images using a
frozen external decoder. Specifically, we reuse the released Stage-1-P
autoencoder from SVG-T2I~\citep{shi2025svgt2i}, whose decoder $D$ was
trained to invert frozen DINOv3 ViT-S/16+ patch features
($D\!:\,\mathbb{R}^{384\times h\times w}\!\to\![-1,1]^{3\times 16h\times 16w}$). At any atlas coordinate $(u,v)$ we evaluate the
composite of Eq.~\eqref{eq:composite}, sample it on a $16\!\times\!16$
canonical grid to produce $z\!\in\!\mathbb{R}^{384\times16\times16}$,
and decode to a $256 \times 256$ image. We then convert the output to grayscale
and apply a 2/98\,\% percentile contrast stretch. The decoder is frozen, is not used during
atlas training, and is shared across datasets.

\begin{figure}[p]
  \centering
  \includegraphics[width=\linewidth]{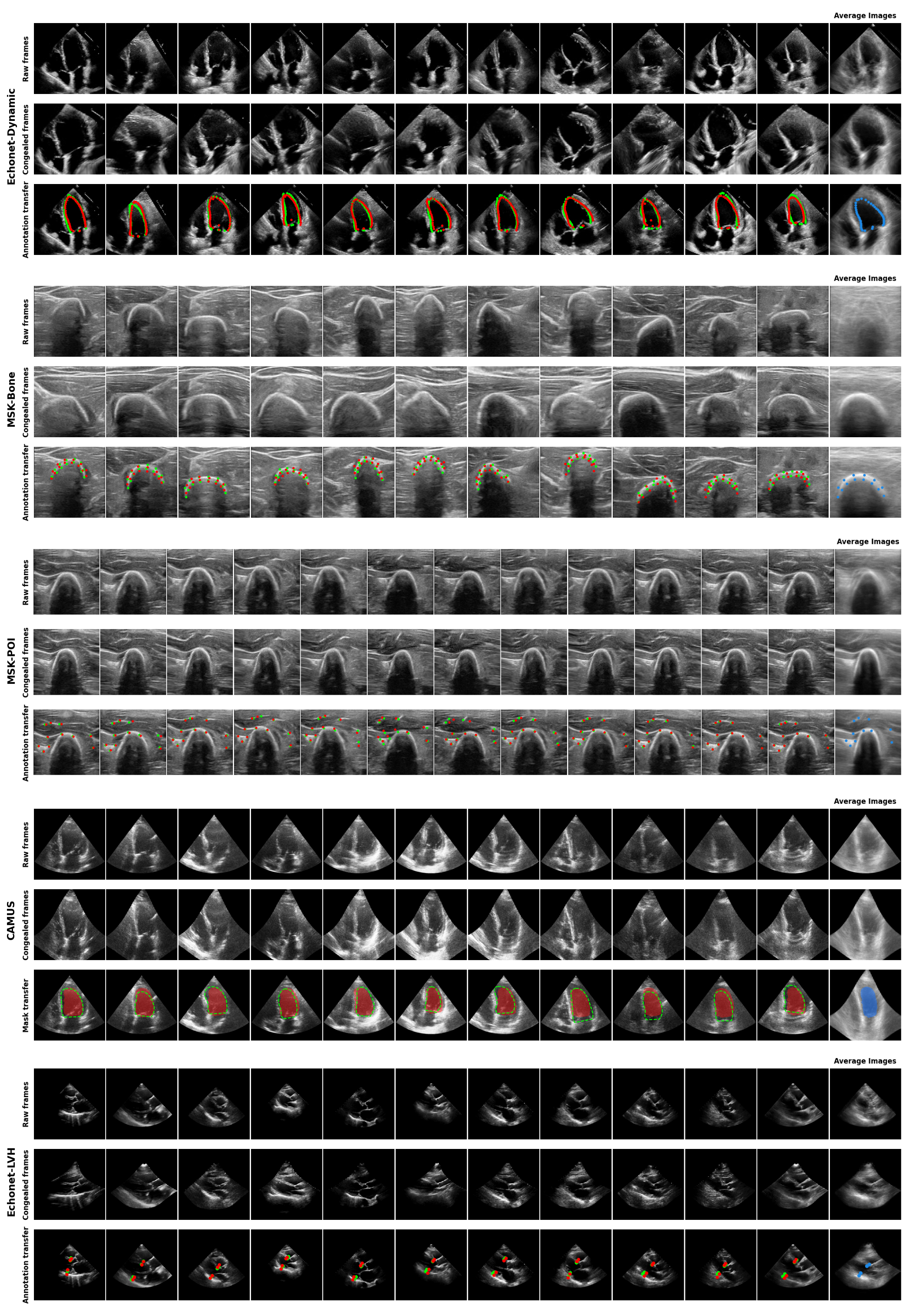}
  \caption{The same atlas pipeline on five ultrasound datasets covering
  cardiac and musculoskeletal anatomies and three annotation types
  (point landmarks, surface-tracking landmarks, segmentation masks).
  Each row-block stacks \emph{raw frames}, \emph{congealed frames}, and \emph{annotation transfer} (GT
  \textcolor{green!70!black}{green}, predictions
  \textcolor{red}{red}), source points in atlas coordinate (\textcolor{blue}{blue}). Within each block, the top and middle rightmost column is the per-row cohort mean in original and common atlas coordinate, the bottom rightmost column is a replicate of middle rightmost column with source points overplayed.
  One atlas per dataset, same training recipe.}
  \label{fig:datasets-hero}
\end{figure}

\section{Single-Shot and Few-Shot Annotation Transfer}
\label{sec:matching}

We benchmark annotation transfer on EchoNet-Dynamic (cardiac, 42-point
left ventricle contour) and MSK-Bone (musculoskeletal, mask-boundary points).
Predictions and ground truth are
linearly rescaled into a fixed $256\!\times\!256$ canvas (TAP-Vid
$\delta_{\mathrm{avg}}$ convention~\citep{doersch2022tapvid}) before
scoring, so EchoNet ($112 \!\times\! 112$) and MSK-Bone ($656\!\times\!496$)
errors are directly comparable. We report mean per-point Euclidean
error (MPE) -- mean / median / std in canvas pixels
(Tables~\ref{tab:paired-mpe}, \ref{tab:ood}). All atlases reported
here use a frozen DINOv3 ViT-B/16 backbone with input images at $448\times448$ resolution.

\paragraph{Qualitative behavior across ultrasound datasets.} Before quantitative evaluation, we assess whether the same architecture
and training recipe produce consistent qualitative behavior across
heterogeneous ultrasound data.
Fig.~\ref{fig:datasets-hero} runs the full pipeline, raw frame,
congealed (inverse-warped) frame and few-shot annotation transfer,
on five datasets~\citep{ouyang2020echonet, duffy2022echonetlvh, leclerc2019camus, pallareslopez2026multimodal} spanning two anatomies (cardiac and musculoskeletal),
multiple acquisition protocols, and three annotation types
(point landmarks, surface-tracking landmarks, and segmentation masks).
For each cohort, the canonical mean resolves into an anatomical
template, and the transferred annotations remain aligned with
ground truth.

\input{tables/T1_paired.tex}

\paragraph{Paired transfer is competitive with strong correspondence
baselines.} Table~\ref{tab:paired-mpe} shows that the proposed atlas is a
strong paired-transfer model despite using a shared canonical representation
rather than direct pairwise matching. Across both ultrasound domains, our atlas
variants achieve stronger overall accuracy and lower dispersion than
matching-based baselines on most paired-transfer summaries, suggesting that
cohort-level canonicalization provides a more reliable correspondence signal
than off-the-shelf image matching. On EchoNet-Dynamic, the full-video model
further shows that unlabeled frames improve the learned atlas without changing
the annotation protocol, while the labeled-only variant remains competitive
when only annotated frames are available.

Neural Congealing highlights the importance of stable population-level atlas
training in medical video. It could only be fit on labeled subsets and
repeatedly failed to converge on EchoNet-Dynamic,
whereas our atlas trains on the full video collection and benefits from
unlabeled frames. On MSK-Bone, Neural Congealing obtains the lowest MPE,
suggesting a complementary inductive bias: standardized acquisition geometry
and clear object boundaries make this dataset well suited to flexible
spatial-transformer warps, while EchoNet-Dynamic has larger variation in
viewpoint, field of view, image quality, and cardiac appearance. In this
setting, the layered atlas with rigidity regularization provides a more robust
correspondence model.

\section{The Atlas Latent Space}
\label{sec:latent}

\begin{figure}[!t]
  \centering
  \includegraphics[width=0.96\linewidth, trim={0pt 8pt 0pt 10pt}, clip]{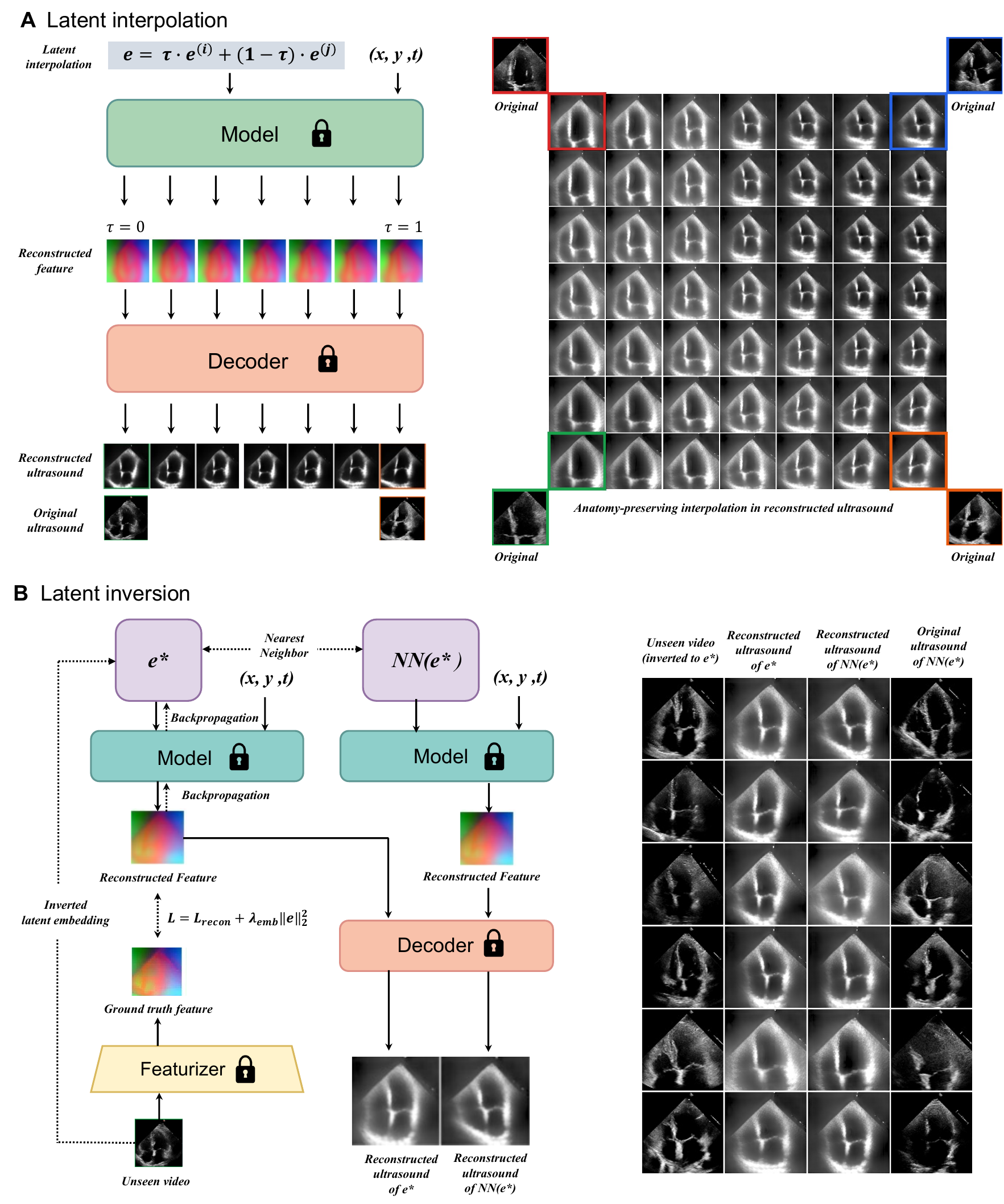}
  \caption{Two operational tests of latent interpretability on the
    EchoNet-Dynamic atlas, both decoded with the frozen off-the-shelf decoder from
    \S\ref{sec:method-decoder}.
    \textbf{(A) Latent interpolation.} Bilinear interpolation among four anchor
    embeddings (red borders) produces a $7\!\times\!7$ grid of decoded frames, where
    chamber geometry, septal thickness, and apex orientation vary smoothly without
    grid-warp or texture artifacts.
    \textbf{(B) Latent inversion.} For an unseen video, a test-time embedding
    $e^{*}$ is optimized to match frozen-featurizer features through the atlas. We
    compare its decoded reconstruction with its nearest neighbor $NN(e^*)$ in
    embedding space, showing both the decoded and original training-set frames.}
  \label{fig:latent}
\end{figure}

The atlas embedding latent admits two operational tests of
interpretability. \emph{Forward}: bilinear interpolation between
known embeddings should decode through the frozen atlas to smooth,
anatomically plausible intermediate frames. \emph{Inverse}: an unseen
frame should be recoverable by optimizing only its embedding while
keeping the atlas fixed. We evaluate both tests in Fig.~\ref{fig:latent}.

\subsection{Anatomy-preserving interpolation}
\label{sec:latent-interp}

Fig.~\ref{fig:latent}A places four anchor videos at the corners of a
$7\!\times\!7$ grid and fills the interior by bilinearly interpolating their
atlas embeddings, $e \!=\! \tau\,e^{(i)} \!+\! (1{-}\tau)\,e^{(j)}$ along each
axis. Each interpolated code is passed through the frozen mapping and atlas
heads to reconstruct a feature grid, then decoded with the same frozen SVG-T2I
decoder described in \S\ref{sec:method-decoder}. No part of the pipeline is retrained:
the model and decoder remain fixed, and only the embedding varies. Across the
grid, chamber geometry, septal thickness, and apex orientation change smoothly
between anchors, while intermediate frames remain anatomically plausible and
avoid the texture or grid-warp artifacts typical of pixel-space interpolation.
Because the decoder was trained on natural images and never adapted to
ultrasound, realistic decoded frames suggest that the atlas embedding space is
locally smooth: interpolated codes remain near the manifold of renderable DINOv3
feature grids.

\subsection{Latent inversion of unseen frames}
\label{sec:latent-invert}

A practical atlas must transfer annotations to videos not seen during atlas
training. We evaluate this setting on held-out EchoNet videos and a
leave-one-out sweep on MSK-Bone (Table~\ref{tab:ood}). For each held-out video,
we freeze the atlas, decoder, and featurizer, and optimize only a new embedding
$e^{*}$ so that the atlas-reconstructed feature grid matches the frozen
featurizer features of the input
(Fig.~\ref{fig:latent}B), using
$\mathcal{L}=\mathcal{L}_{\mathrm{recon}}+
\lambda_{\mathrm{emb}}\|e\|_2^2$ for roughly 200 Adam steps
($\sim$10\,s on a single RTX 4090). We then transfer annotations through the
recovered embedding using $K\!\in\!\{2,5,10\}$ source videos.

Latent inversion achieves accuracy comparable to the in-atlas setting, where
each video's embedding was learned jointly during atlas training, and performance
saturates after only a few support videos. Fig.~\ref{fig:latent}B provides a
visual check that $e^{*}$ does more than retrieve its closest training example:
for several unseen frames, the inverted reconstruction is closer to the input
than both the decoded nearest neighbor $\mathrm{NN}(e^{*})$ in latent space and
the neighbor's original training frame. This suggests that gradient-based
inversion reaches meaningful new points in the atlas latent space rather than
copying existing cohort members. Tshese results show that the latent space is both locally
smooth and expressive enough to represent held-out variation.

\input{tables/T2_ood.tex}

\section{Ablations}
\label{sec:ablations}

\begin{table}[t]
\small
\centering
\setlength{\tabcolsep}{4pt}
\caption{Component ablations. Each row removes one component from
the full model and re-runs the paired-transfer evaluation of
Table~\ref{tab:paired-mpe}; all other settings fixed. Mean MPE in
pixels on a $256\!\times\!256$ canvas (lower is better);
\textbf{bold} = best entry, \underline{underline} = second best.}
\label{tab:ablation}
\begin{tabular}{llcc}
\toprule
Family & Variant & EchoNet-Dynamic & MSK-Bone \\

\midrule
-- & \textbf{Full model}                  & \textbf{17.74}    & \underline{13.74} \\
\midrule
\multirow{5}{*}{Regularizer}
 & $-\,\mathcal{L}_{\mathrm{rig}}$ (rigidity)             & 18.27             & 13.75             \\
 & $-\,\mathcal{L}_{\mathrm{d}}$ (dynamic-atlas smoothness) & 17.88           & \textbf{13.50}    \\
 & $-\,\mathcal{L}_{\mathrm{s}}$ (static-atlas smoothness)  & \underline{17.74} & 13.75           \\
 & $-\,\mathcal{L}_{\mathrm{div}}$ (diversity)            & 17.79             & 13.77             \\
 & $-\,\mathcal{L}_{\mathrm{anc}}$ (anchor at $e{=}0$)    & 37.71             & 47.66             \\
\midrule
\multirow{2}{*}{Architecture}
 & $-$ background branch                 & 18.06             & 16.48             \\
 & $-$ uncertainty ($\beta$) head        & 21.01             & 17.51             \\
\bottomrule
\end{tabular}
\end{table}

The pattern is consistent across anatomies
(Table~\ref{tab:ablation}). Removing the anchor
$\mathcal{L}_{\mathrm{anc}}$ catastrophically degrades transfer accuracy on
both datasets, producing the largest jump in the table: without a fixed
canonical chart at $e{=}\mathbf{0}$, the latent space can drift and inverse
search lands in the wrong branch. The architectural removals are next most
impactful: removing the uncertainty ($\beta$) head or static-atlas branch
notably hurts MSK-Bone, while effect is smaller on EchoNet. Removing
the regularizers
$\mathcal{L}_{\mathrm{rig}}$,
$\mathcal{L}_{\mathrm{d}}$,
$\mathcal{L}_{\mathrm{s}}$, and
$\mathcal{L}_{\mathrm{div}}$
changes mean MPE only marginally in isolation, but they stabilize training.

\section{Discussion and Conclusion}
\label{sec:discussion}

We introduced a cohort-scale neural atlas for ultrasound video that factorizes
each sequence into shared canonical structure, per-video deformation, and
per-video identity. Implemented as a compact four-head SIREN with
NeRF-W-style uncertainty and GLO embeddings, the model trains jointly across
hundreds of videos and thousands of frames in minutes on a single consumer GPU.
This factorization supports single-shot and few-shot atlas-mediated annotation
transfer, outperforming strong dense-correspondence baselines, while exposing an
interpretable latent space through interpolation, test-time inversion, and
off-the-shelf DINOv3-decoder visualization. Unlike generic feature matching, the
learned factors are directly manipulable: canonical coordinates can be queried,
deformations inverted, and identity embeddings optimized or interpolated. These
properties make the atlas a reusable ultrasound-video representation for
annotation transfer, biomarker estimation, quality control, longitudinal and
cross-modality analysis, and privacy-conscious multi-site learning.

\paragraph{Limitations.} Annotation transfer currently requires test-time
inverse search and local optimization
(\S\ref{sec:method-transfer}), rather than feed-forward prediction. Future work
could amortize this step with a lightweight encoder. Quantitative validation is
limited to two ultrasound anatomies: cardiac views and upper-limb
musculoskeletal scans. Finally, rigidity regularization stabilizes population
alignment but can underfit highly local non-rigid variation, as in MSK-Bone.
Staged rigid-to-deformable registration may combine both regimes.

\bibliographystyle{plainnat}
\bibliography{refs}

\end{document}

%% file: tables/T1_paired.tex
\begin{table}[t]
\small
\centering
\caption{Paired video annotation transfer (MPE on a 256$\times$256 canvas, in pixels). Lower is better. \textbf{Best} and \underline{second-best} are highlighted.}
\label{tab:paired-mpe}
\begin{tabular}{lcccccc}
\toprule
 & \multicolumn{3}{c}{EchoNet-Dynamic (196 pairs)} & \multicolumn{3}{c}{MSK-Bone (380 pairs)} \\
\cmidrule(lr){2-4}\cmidrule(lr){5-7}
Method & mean & median & std & mean & median & std \\
\midrule
\multicolumn{7}{l}{\emph{Matching-based}} \\
DIFT~\citep{tang2023dift}             & 41.89 & 31.16 & 34.46 & 20.07 & 11.69 & 26.81 \\
MATCHA~\citep{xue2025matcha}          & 28.01 & 20.22 & 24.18 & 16.70 & 11.01 & 19.75 \\
RoMa-indoor~\citep{edstedt2024roma}   & 30.90 & 20.91 & 30.47 & 47.21 & 24.78 & 48.32 \\
RoMa-outdoor~\citep{edstedt2024roma}  & 25.86 & 19.66 & 21.87 & 46.27 & 27.89 & 43.85 \\
\midrule
\multicolumn{7}{l}{\emph{Atlas-based}} \\
\makecell[l]{Neural Congealing*~\citep{ofriamar2023neural} \\ (labeled only)} & - & - & - & \textbf{9.83} & \textbf{8.24} & \textbf{6.77} \\
Ours (labeled only) & \textbf{17.72} & \underline{14.55}    & \textbf{14.44}    & \underline{11.89} & \underline{10.18} & \underline{7.68} \\
Ours (full video)   & \underline{17.74}    & \textbf{14.42}    & \underline{14.91} & 13.74 & 12.09 & 8.54 \\

\bottomrule
\multicolumn{5}{l}{\footnotesize $^*$Failed to converge on EchoNet-Dynamic dataset in atlas training.}
\end{tabular}
\end{table}

%% file: tables/T2_ood.tex
\begin{table}[t]
\small
\centering
\caption{Out-of-atlas-training annotation transfer via latent inversion
(\S\ref{sec:latent-invert}, Fig.~\ref{fig:latent}B). For each held-out video,
we freeze the atlas and decoder, optimize a test-time embedding $e^{*}$, and
transfer annotations from $K$ source videos. \emph{In atlas-training} reports the
oracle jointly learned embedding. \emph{Out of atlas-training} reports latent
inversion. Cells show MPE mean / median / std on a $256\!\times\!256$ canvas
(lower is better). Per metric, \textbf{best} and \underline{second-best} are
highlighted within each column.}
\label{tab:ood}
\setlength{\tabcolsep}{4pt}
\begin{tabular}{ccccc}
\toprule
 & \multicolumn{4}{c}{MPE mean / median / std $\downarrow$ (canvas\,px)} \\
 \cmidrule(lr){2-5}
 & \multicolumn{2}{c}{EchoNet-Dynamic} & \multicolumn{2}{c}{MSK-Bone} \\
 \cmidrule(lr){2-3}\cmidrule(lr){4-5}
 $K$ & in atlas-training & out of atlas-training & in atlas-training & out of atlas-training \\
\midrule
 2  & 14.35 / 12.54 / 10.94
    & 13.90 / 11.69 / 11.69
    & 11.74 / 10.55 / 6.95
    & 13.59 / 12.25 / 8.12 \\
 5  & \underline{13.43} / \underline{11.49} / \underline{10.83}
    & \textbf{12.85} / \textbf{10.61} / \underline{11.66}
    & \underline{10.55} / \textbf{9.07} / \underline{6.82}
    & \underline{12.89} / \underline{11.42} / \underline{7.91} \\
 10 & \textbf{13.30} / \textbf{11.48} / \textbf{10.70}
    & \underline{12.94} / \underline{10.64} / \textbf{11.63}
    & \textbf{10.43} / \underline{9.18} / \textbf{6.48}
    & \textbf{12.67} / \textbf{11.27} / \textbf{7.51} \\
\bottomrule
\end{tabular}
\end{table}

%% file: main.bbl
\begin{thebibliography}{35}
\providecommand{\natexlab}[1]{#1}
\providecommand{\url}[1]{\texttt{#1}}
\expandafter\ifx\csname urlstyle\endcsname\relax
  \providecommand{\doi}[1]{doi: #1}\else
  \providecommand{\doi}{doi: \begingroup \urlstyle{rm}\Url}\fi

\bibitem[Avants et~al.(2008)Avants, Epstein, Grossman, and Gee]{avants2008syn}
Brian~B. Avants, Charles~L. Epstein, Murray Grossman, and James~C. Gee.
\newblock Symmetric diffeomorphic image registration with cross-correlation:
  Evaluating automated labeling of elderly and neurodegenerative brain.
\newblock \emph{Medical Image Analysis}, 12\penalty0 (1):\penalty0 26--41,
  2008.
\newblock \doi{10.1016/j.media.2007.06.004}.

\bibitem[Balakrishnan et~al.(2019)Balakrishnan, Zhao, Sabuncu, Guttag, and
  Dalca]{balakrishnan2019voxelmorph}
Guha Balakrishnan, Amy Zhao, Mert~R. Sabuncu, John Guttag, and Adrian~V. Dalca.
\newblock {VoxelMorph}: A learning framework for deformable medical image
  registration.
\newblock \emph{IEEE Transactions on Medical Imaging}, 38\penalty0
  (8):\penalty0 1788--1800, 2019.
\newblock \doi{10.1109/TMI.2019.2897538}.

\bibitem[Bojanowski et~al.(2018)Bojanowski, Joulin, Lopez-Paz, and Szlam]{glo}
Piotr Bojanowski, Armand Joulin, David Lopez-Paz, and Arthur Szlam.
\newblock Optimizing the latent space of generative networks.
\newblock In \emph{Proceedings of the International Conference on Machine
  Learning}, 2018.

\bibitem[Cabezas et~al.(2011)Cabezas, Oliver, Llad{\'o}, Freixenet, and
  Cuadra]{cabezas2011atlasseg}
Mariano Cabezas, Arnau Oliver, Xavier Llad{\'o}, Jordi Freixenet, and
  Meritxell~Bach Cuadra.
\newblock A review of atlas-based segmentation for magnetic resonance brain
  images.
\newblock \emph{Computer Methods and Programs in Biomedicine}, 104\penalty0
  (3):\penalty0 e158--e177, 2011.

\bibitem[Caron et~al.(2021)Caron, Touvron, Misra, J{\'e}gou, Mairal,
  Bojanowski, and Joulin]{caron2021emerging}
Mathilde Caron, Hugo Touvron, Ishan Misra, Herv{\'e} J{\'e}gou, Julien Mairal,
  Piotr Bojanowski, and Armand Joulin.
\newblock Emerging properties in self-supervised vision transformers.
\newblock In \emph{Proceedings of the IEEE/CVF International Conference on
  Computer Vision}, pages 9630--9640, 2021.

\bibitem[Collins et~al.(1994)Collins, Neelin, Peters, and
  Evans]{collins1994mni}
D.~Louis Collins, Peter Neelin, Terry~M. Peters, and Alan~C. Evans.
\newblock Automatic {3-D} intersubject registration of {MR} volumetric data in
  standardized talairach space.
\newblock \emph{Journal of Computer Assisted Tomography}, 18\penalty0
  (2):\penalty0 192--205, 1994.

\bibitem[Deng et~al.(2024)Deng, Wu, Zeng, and Qin]{deng2024memsam}
Xiaolong Deng, Huisi Wu, Runhao Zeng, and Jing Qin.
\newblock {MemSAM}: Taming segment anything model for echocardiography video
  segmentation.
\newblock In \emph{Proceedings of the IEEE/CVF Conference on Computer Vision
  and Pattern Recognition}, 2024.

\bibitem[Doersch et~al.(2022)Doersch, Gupta, Markeeva, Recasens, Smaira, Aytar,
  Carreira, Zisserman, and Yang]{doersch2022tapvid}
Carl Doersch, Ankush Gupta, Larisa Markeeva, Adri{\`a} Recasens, Lucas Smaira,
  Yusuf Aytar, Jo{\~a}o Carreira, Andrew Zisserman, and Yi~Yang.
\newblock {TAP-Vid}: A benchmark for tracking any point in a video.
\newblock In \emph{Advances in Neural Information Processing Systems}, 2022.

\bibitem[Duffy et~al.(2022)Duffy, Cheng, Yuan, He, Kwan, Shun-Shin, Alexander,
  Ebinger, Lungren, Rader, Liang, Schnittger, Ashley, Zou, Patel, Witteles,
  Cheng, and Ouyang]{duffy2022echonetlvh}
Grant Duffy, Paul~P. Cheng, Neal Yuan, Bryan He, Alan~C. Kwan, Matthew~J.
  Shun-Shin, Kevin~M. Alexander, Joseph Ebinger, Matthew~P. Lungren, Florian
  Rader, David~H. Liang, Ingela Schnittger, Euan~A. Ashley, James~Y. Zou,
  Jignesh Patel, Ronald Witteles, Susan Cheng, and David Ouyang.
\newblock High-throughput precision phenotyping of left ventricular hypertrophy
  with cardiovascular deep learning.
\newblock \emph{JAMA Cardiology}, 7\penalty0 (4):\penalty0 386--395, 2022.
\newblock \doi{10.1001/jamacardio.2021.6059}.

\bibitem[Edstedt et~al.(2024)Edstedt, Sun, Bokman, Wadenb{\"a}ck, and
  Felsberg]{edstedt2024roma}
Johan Edstedt, Qiyu Sun, Georg Bokman, M{\aa}rten Wadenb{\"a}ck, and Michael
  Felsberg.
\newblock {RoMa}: Robust dense feature matching.
\newblock In \emph{Proceedings of the IEEE/CVF Conference on Computer Vision
  and Pattern Recognition}, 2024.

\bibitem[Evans et~al.(1993)Evans, Collins, Mills, Brown, Kelly, and
  Peters]{evans1993mni}
Alan~C. Evans, D.~Louis Collins, S.R. Mills, E.D. Brown, R.L. Kelly, and
  Terry~M. Peters.
\newblock {3D} statistical neuroanatomical models from 305 {MRI} volumes.
\newblock \emph{IEEE Nuclear Science Symposium and Medical Imaging Conference},
  1993.

\bibitem[Fischl(2012)]{fischl2012freesurfer}
Bruce Fischl.
\newblock {FreeSurfer}.
\newblock \emph{NeuroImage}, 62\penalty0 (2):\penalty0 774--781, 2012.

\bibitem[Fu et~al.(2024)Fu, Hamilton, Brandt, Feldmann, Zhang, and
  Freeman]{fu2024featup}
Stephanie Fu, Mark Hamilton, Laura Brandt, Axel Feldmann, Zhoutong Zhang, and
  William~T. Freeman.
\newblock {FeatUp}: A model-agnostic framework for features at any resolution.
\newblock In \emph{International Conference on Learning Representations}, 2024.

\bibitem[Gupta et~al.(2023)Gupta, Jampani, Esteves, Shrivastava, Makadia,
  Snavely, and Kar]{gupta2023asic}
Kamal Gupta, Varun Jampani, Carlos Esteves, Abhinav Shrivastava, Ameesh
  Makadia, Noah Snavely, and Abhishek Kar.
\newblock {ASIC}: Aligning sparse in-the-wild image collections.
\newblock In \emph{Proceedings of the IEEE/CVF International Conference on
  Computer Vision}, 2023.

\bibitem[Kasten et~al.(2021)Kasten, Ofri, Wang, and Dekel]{kasten2021layered}
Yoni Kasten, Dolev Ofri, Oliver Wang, and Tali Dekel.
\newblock Layered neural atlases for consistent video editing.
\newblock \emph{ACM Transactions on Graphics}, 40\penalty0 (6), 2021.
\newblock \doi{10.1145/3478513.3480546}.

\bibitem[Learned-Miller(2006)]{learnedmiller2006congealing}
Erik~G. Learned-Miller.
\newblock Data driven image models through continuous joint alignment.
\newblock \emph{IEEE Transactions on Pattern Analysis and Machine
  Intelligence}, 28\penalty0 (2):\penalty0 236--250, 2006.
\newblock \doi{10.1109/TPAMI.2006.34}.

\bibitem[Leclerc et~al.(2019)Leclerc, Smistad, Pedrosa, {\O}stvik, Cervenansky,
  Espinosa, Espeland, Berg, Jodoin, Grenier, Lartizien, D'hooge, L{\o}vstakken,
  and Bernard]{leclerc2019camus}
Sarah Leclerc, Erik Smistad, Jo{\~a}o Pedrosa, Andreas {\O}stvik, Frederic
  Cervenansky, Florian Espinosa, Torvald Espeland, Erik Andreas~Rye Berg,
  Pierre-Marc Jodoin, Thomas Grenier, Carole Lartizien, Jan D'hooge, Lasse
  L{\o}vstakken, and Olivier Bernard.
\newblock Deep learning for segmentation using an open large-scale dataset in
  2d echocardiography.
\newblock \emph{IEEE Transactions on Medical Imaging}, 38\penalty0
  (9):\penalty0 2198--2210, 2019.
\newblock \doi{10.1109/TMI.2019.2900516}.

\bibitem[Martin-Brualla et~al.(2021)Martin-Brualla, Radwan, Sajjadi, Barron,
  Dosovitskiy, and Duckworth]{nerfw}
Ricardo Martin-Brualla, Noha Radwan, Mehdi~S.M. Sajjadi, Jonathan~T. Barron,
  Alexey Dosovitskiy, and Daniel Duckworth.
\newblock {NeRF} in the wild: Neural radiance fields for unconstrained photo
  collections.
\newblock In \emph{Proceedings of the IEEE/CVF Conference on Computer Vision
  and Pattern Recognition}, 2021.

\bibitem[Mildenhall et~al.(2020)Mildenhall, Srinivasan, Tancik, Barron,
  Ramamoorthi, and Ng]{mildenhall2020nerf}
Ben Mildenhall, Pratul~P. Srinivasan, Matthew Tancik, Jonathan~T. Barron, Ravi
  Ramamoorthi, and Ren Ng.
\newblock {NeRF}: Representing scenes as neural radiance fields for view
  synthesis.
\newblock In \emph{Proceedings of the European Conference on Computer Vision},
  2020.

\bibitem[Namburi et~al.(2025)Namburi, Pallar{\`e}s-L{\'o}pez, Folgado,
  Magana-Salgado, Rosendorf, Ryu, Kappacher, Gamboa, Anthony, and
  Daniel]{namburi2025efficient}
Praneeth Namburi, Roger Pallar{\`e}s-L{\'o}pez, Duarte Folgado, Uriel
  Magana-Salgado, Jessica Rosendorf, Enya Ryu, Armin Kappacher, Hugo Gamboa,
  Brian~W. Anthony, and Luca Daniel.
\newblock Efficient elastic tissue motions indicate general motor skill.
\newblock \emph{Scientific Reports}, 15\penalty0 (1):\penalty0 36532, 2025.
\newblock \doi{10.1038/s41598-025-17092-0}.
\newblock URL \url{https://www.nature.com/articles/s41598-025-17092-0}.

\bibitem[Ofri-Amar et~al.(2023)Ofri-Amar, Geyer, Kasten, and
  Dekel]{ofriamar2023neural}
Dolev Ofri-Amar, Michal Geyer, Yoni Kasten, and Tali Dekel.
\newblock Neural congealing: Aligning images to a joint semantic atlas.
\newblock In \emph{Proceedings of the IEEE/CVF Conference on Computer Vision
  and Pattern Recognition}, 2023.

\bibitem[Oquab et~al.(2024)Oquab, Darcet, Moutakanni, Vo, Szafraniec, Khalidov,
  Fernandez, Haziza, Massa, El-Nouby, Assran, Ballas, Galuba, Howes, Huang, Li,
  Misra, Rabbat, Sharma, Synnaeve, Xu, J{\'e}gou, Mairal, Labatut, Joulin, and
  Bojanowski]{oquab2023dinov2}
Maxime Oquab, Timoth{\'e}e Darcet, Th{\'e}o Moutakanni, Huy~V. Vo, Marc
  Szafraniec, Vasil Khalidov, Pierre Fernandez, Daniel Haziza, Francisco Massa,
  Alaaeldin El-Nouby, Mahmoud Assran, Nicolas Ballas, Wojciech Galuba, Russell
  Howes, Po-Yao Huang, Shang-Wen Li, Ishan Misra, Michael Rabbat, Vasu Sharma,
  Gabriel Synnaeve, Hu~Xu, Herv{\'e} J{\'e}gou, Julien Mairal, Patrick Labatut,
  Armand Joulin, and Piotr Bojanowski.
\newblock {DINOv2}: Learning robust visual features without supervision.
\newblock \emph{Transactions on Machine Learning Research}, 2024.

\bibitem[Ouyang et~al.(2020)Ouyang, He, Ghorbani, Yuan, Ebinger, Langlotz,
  Heidenreich, Harrington, Liang, Ashley, and Zou]{ouyang2020echonet}
David Ouyang, Bryan He, Amirata Ghorbani, Neal Yuan, Joseph Ebinger, Curt~P.
  Langlotz, Paul~A. Heidenreich, Robert~A. Harrington, David~H. Liang, Euan~A.
  Ashley, and James~Y. Zou.
\newblock Video-based {AI} for beat-to-beat assessment of cardiac function.
\newblock \emph{Nature}, 580:\penalty0 252--256, 2020.
\newblock \doi{10.1038/s41586-020-2145-8}.

\bibitem[Ouyang et~al.(2024)Ouyang, Wang, Xiao, Bai, Zhang, Zheng, Zhou, Chen,
  and Shen]{ouyang2024codef}
Hao Ouyang, Qiuyu Wang, Yuxi Xiao, Qingyan Bai, Juntao Zhang, Kecheng Zheng,
  Xiaowei Zhou, Qifeng Chen, and Yujun Shen.
\newblock {CoDeF}: Content deformation fields for temporally consistent video
  processing.
\newblock In \emph{Proceedings of the IEEE/CVF Conference on Computer Vision
  and Pattern Recognition}, 2024.

\bibitem[Pallar{\`e}s-L{\'o}pez et~al.(2026)Pallar{\`e}s-L{\'o}pez, Folgado,
  Magana-Salgado, Rosendorf, Ryu, Feigin-Almon, Gamboa, Daniel, Anthony, and
  Namburi]{pallareslopez2026multimodal}
Roger Pallar{\`e}s-L{\'o}pez, Duarte Folgado, Uriel Magana-Salgado, Jessica
  Rosendorf, Enya Ryu, Micha Feigin-Almon, Hugo Gamboa, Luca Daniel, Brian~W.
  Anthony, and Praneeth Namburi.
\newblock A multimodal biomechanics dataset with synchronized kinematics and
  internal tissue motions during reaching.
\newblock \emph{Scientific Data}, 2026.
\newblock \doi{10.1038/s41597-026-07019-3}.

\bibitem[Peebles et~al.(2022)Peebles, Zhu, Zhang, Torralba, Efros, and
  Shechtman]{peebles2022gangealing}
William Peebles, Jun-Yan Zhu, Richard Zhang, Antonio Torralba, Alexei Efros,
  and Eli Shechtman.
\newblock {GAN}-supervised dense visual alignment.
\newblock In \emph{Proceedings of the IEEE/CVF Conference on Computer Vision
  and Pattern Recognition}, 2022.

\bibitem[Shi et~al.(2025)Shi, Wang, Zhang, Zheng, Zeng, Yuan, Wu, Zhang, Yang,
  Wang, Wan, Gai, Zhou, and Lu]{shi2025svgt2i}
Minglei Shi, Haolin Wang, Borui Zhang, Wenzhao Zheng, Bohan Zeng, Ziyang Yuan,
  Xiaoshi Wu, Yuanxing Zhang, Huan Yang, Xintao Wang, Pengfei Wan, Kun Gai, Jie
  Zhou, and Jiwen Lu.
\newblock {SVG-T2I}: Scaling up text-to-image latent diffusion model without
  variational autoencoder.
\newblock \emph{arXiv preprint arXiv:2512.11749}, 2025.

\bibitem[Sim{\'e}oni et~al.(2025)Sim{\'e}oni, Vo, Seitzer, Baldassarre, Oquab,
  Jose, Khalidov, Szafraniec, Yi, Ramamonjisoa, Massa, Haziza, Wehrstedt, Wang,
  Darcet, Moutakanni, Sentana, Roberts, Vedaldi, Tolan, Brandt, Couprie,
  Mairal, J{\'e}gou, Labatut, and Bojanowski]{simeoni2025dinov3}
Oriane Sim{\'e}oni, Huy~V. Vo, Maximilian Seitzer, Federico Baldassarre, Maxime
  Oquab, Cijo Jose, Vasil Khalidov, Marc Szafraniec, Seungeun Yi, Micha{\"e}l
  Ramamonjisoa, Francisco Massa, Daniel Haziza, Luca Wehrstedt, Jianyuan Wang,
  Timoth{\'e}e Darcet, Th{\'e}o Moutakanni, Leonel Sentana, Claire Roberts,
  Andrea Vedaldi, Jamie Tolan, John Brandt, Camille Couprie, Julien Mairal,
  Herv{\'e} J{\'e}gou, Patrick Labatut, and Piotr Bojanowski.
\newblock {DINOv3}, 2025.
\newblock URL \url{https://arxiv.org/abs/2508.10104}.

\bibitem[Sitzmann et~al.(2020)Sitzmann, Martel, Bergman, Lindell, and
  Wetzstein]{sitzmann2020siren}
Vincent Sitzmann, Julien~N.P. Martel, Alexander~W. Bergman, David~B. Lindell,
  and Gordon Wetzstein.
\newblock Implicit neural representations with periodic activation functions.
\newblock In \emph{Advances in Neural Information Processing Systems}, 2020.

\bibitem[Talairach and Tournoux(1988)]{talairach1988atlas}
Jean Talairach and Pierre Tournoux.
\newblock \emph{Co-Planar Stereotaxic Atlas of the Human Brain: {3-Dimensional}
  Proportional System: An Approach to Cerebral Imaging}.
\newblock Thieme, 1988.

\bibitem[Tancik et~al.(2020)Tancik, Srinivasan, Mildenhall, Fridovich-Keil,
  Raghavan, Singhal, Ramamoorthi, Barron, and Ng]{tancik2020fourier}
Matthew Tancik, Pratul~P. Srinivasan, Ben Mildenhall, Sara Fridovich-Keil,
  Nithin Raghavan, Utkarsh Singhal, Ravi Ramamoorthi, Jonathan~T. Barron, and
  Ren Ng.
\newblock Fourier features let networks learn high frequency functions in low
  dimensional domains.
\newblock In \emph{Advances in Neural Information Processing Systems},
  volume~33, pages 7537--7547, 2020.

\bibitem[Tang et~al.(2023)Tang, Jia, Wang, Phoo, and Hariharan]{tang2023dift}
Luming Tang, Menglin Jia, Qianqian Wang, Cheng~Perng Phoo, and Bharath
  Hariharan.
\newblock Emergent correspondence from image diffusion.
\newblock In \emph{Advances in Neural Information Processing Systems}, 2023.

\bibitem[Xie et~al.(2022)Xie, Takikawa, Saito, Litany, Yan, Khan, Tombari,
  Tompkin, Sitzmann, and Sridhar]{xie2021neuralfields}
Yiheng Xie, Towaki Takikawa, Shunsuke Saito, Or~Litany, Shiqin Yan, Numair
  Khan, Federico Tombari, James Tompkin, Vincent Sitzmann, and Srinath Sridhar.
\newblock Neural fields in visual computing and beyond.
\newblock \emph{Computer Graphics Forum}, 41\penalty0 (2):\penalty0 641--676,
  2022.
\newblock \doi{10.1111/cgf.14505}.

\bibitem[Xue et~al.(2025)Xue, Elflein, Leal-Taix{\'e}, and Zhou]{xue2025matcha}
Fei Xue, Sven Elflein, Laura Leal-Taix{\'e}, and Qunjie Zhou.
\newblock {MATCHA}: Towards matching anything.
\newblock In \emph{Proceedings of the IEEE/CVF Conference on Computer Vision
  and Pattern Recognition}, 2025.

\bibitem[Zhang et~al.(2024)Zhang, Li, Raj, Engelhardt, Li, Hou, Wu, and
  Jampani]{zhang2024congealing3d}
Yunzhi Zhang, Zizhang Li, Amit Raj, Andreas Engelhardt, Yuanzhen Li, Tingbo
  Hou, Jiajun Wu, and Varun Jampani.
\newblock {3D} congealing: {3D}-aware image alignment in the wild.
\newblock In \emph{Proceedings of the European Conference on Computer Vision},
  2024.

\end{thebibliography}
